\documentclass[conference]{IEEEtran}
\IEEEoverridecommandlockouts
\usepackage{cite}
\usepackage{amsmath,amssymb,amsfonts}
\usepackage{algorithmic}
\usepackage{graphicx}
\usepackage{textcomp}
\usepackage{xcolor}
\usepackage{physics}
\usepackage{hyperref}

\usepackage{url}            % simple URL typesetting
\usepackage{booktabs}       % professional-quality tables
\usepackage{nicefrac}       % compact symbols for 1/2, etc.
\usepackage{microtype}      % microtypography
\usepackage{physics}
\usepackage{comment}
\usepackage{mathrsfs}
\usepackage[makeroom]{cancel}
\usepackage{clrscode}
\usepackage{tikz}
\usepackage{multirow}
\usepackage{multicol}
\usepackage{caption}
\usepackage{subcaption}

\newcommand{\reals}[1]{\mathbb{R}^{#1}}
\newcommand{\myforall}[0]{\ \forall \ }

\newcommand{\bigoh}[1]{\mathcal{O}(#1)}
\newcommand{\pid}[0]{\mathit{PID}}

\def\BibTeX{{\rm B\kern-.05em{\sc i\kern-.025em b}\kern-.08em
    T\kern-.1667em\lower.7ex\hbox{E}\kern-.125emX}}
\begin{document}

\title{Encoding Involutory Invariances in Neural Networks\\
}

\author{\IEEEauthorblockN{1\textsuperscript{st} Anwesh Bhattacharya}
\IEEEauthorblockA{\textit{Dept. of Physics \& CSIS} \\
\textit{BITS-Pilani, Pilani}\\
Rajasthan, India \\
f2016590@pilani.bits-pilani.ac.in}
\and
\IEEEauthorblockN{2\textsuperscript{nd} Marios Mattheakis}
\IEEEauthorblockA{\textit{Institute for Applied Computational Science} \\
\textit{Harvard University}\\
Cambridge, Massachusetts, United States \\
mariosmat@seas.harvard.edu
}
\and
\IEEEauthorblockN{3\textsuperscript{rd} Pavlos Protopapas}
\IEEEauthorblockA{\textit{Institute for Applied Computational Science} \\
\textit{Harvard University}\\
Cambridge, Massachusetts, United States \\
pavlos@seas.harvard.edu
}
}

\maketitle

\begin{abstract}
    In certain situations, neural networks are trained upon data that obey underlying symmetries. However, the predictions do not respect the symmetries \emph{exactly} unless embedded in the network structure. In this work, we introduce architectures that embed a special kind of symmetry namely, invariance with respect to involutory linear/affine transformations up to parity $p=\pm 1$. We provide rigorous theorems to show that the proposed network ensures such an invariance and present qualitative arguments for a special universal approximation theorem. An adaption of our techniques to CNN tasks for datasets with inherent horizontal/vertical reflection symmetry is demonstrated. Extensive experiments indicate that the proposed model outperforms baseline feed-forward and physics-informed neural networks while identically respecting the underlying symmetry.
\end{abstract}

\begin{IEEEkeywords}
neural networks, symmetries, invariances, universal approximation
\end{IEEEkeywords}

\section{Introduction}

There has been notable effort in the direction of encoding symmetries in Neural Networks (NN). This is particularly important as various real-life tasks have inherent symmetries associated with them. Moreover, the act of encoding such symmetries makes the behaviour of neural networks for explainable. For instance, the proposed architecture in \cite{mattheakis.symmetry} encodes symmetries in NNs that make it capable of learning even/odd functions. 

\cite{navier.stokes} explores techniques for solving the Navier-Stokes equations while guaranteeing its various symmetries, namely  space/time translation, rotation, reflection, and scaling. \cite{lorentz} builds networks that obey equivariance to Lorenz Transformations \cite{resnick} which is essential in problems centred around particle physics. \cite{nn.art} experiments with feeding NNs art paintings and conduct a principal component analysis  on the weights of the final dense layer of the network. They  found that the distribution of the weights reflect the underlying symmetries of the images.   \cite{beyond.permutation} proposes novel architectures for graph networks that are equivariant to the Euclidean group.

Significant effort has been spent on guaranteeing continuous group symmetries such as \cite{esteves} which computes rotation invariant feature maps for 3D CAD models. \cite{cohen} have developed NNs with generalized group-theoretic features such as G-convolutions. A theoretical analysis of the connection between group convolutions and equivariance with respect to actions of a group can be found in \cite{kondor}. 

The aforementioned studies are significant in their own right to guaranteeing general invariance/equivariances, however, they use sophisticated networks and require in-depth knowledge of group theory. In this work, we present a simplistic  exploit of the standard Feed-Forward Neural Network (FFNN) \cite{mlp} that guarantees a special type of symmetry known as \emph{involutory invariance}, a  symmetry that has not be studied yet in the context of deep learning. Involutory transformations encode several physical relevant symmetries that are present in various academic (Physics datasets) and practical scenarios (Image Classification).

\subsection{Why Involutory Transformations?}

An involutory transformation is one that is its own inverse. Some physically relevant symmetries\footnote{
Throughout this paper, we have used the word \emph{symmetry} and \emph{invariance} interchangably. The two words essentially refer to the same concept.
} fall into the category of involutory transformations such as reflection about a plane, or inversion about a point. These transformations enjoy certain properties which make it suitable for NNs to learn functions that are invariant to such transformations. We hereon represent an involutory transformation as the matrix $A$.

\textbf{Example:} The reflection matrix about the XY-plane in 3D, and inversion matrix in 2D would respectively be
\begin{gather*}
    \begin{bmatrix}
    1 & 0 & 0 \\
    0 & 1 & 0 \\
    0 & 0 & -1
    \end{bmatrix} \text{ and }
    \begin{bmatrix}
    -1 & 0 \\
    0 & -1 \\
    \end{bmatrix}. \label{eq:example.matrices}
\end{gather*}
The above two transformations represent involutory \emph{linear} transformation. It is possible to conceive of involutory \emph{affine} transformations as well. Refer to section \ref{sec:affine} for a discussion on this subtle aspect.

\subsection{Organization of the Paper}

We first recapitulate the hub-layered NN  \cite{mattheakis.symmetry} in Section \ref{sec:hub.recap}\ which serves as the basis for this work. Section \ref{sec:san} contains an important result on classes of symmetry-inducing activation functions. Subsequently, the Involutory Partition Theorem (IPT) is presented in Section \ref{sec:iptnet} that endows FFNNs the capability of learning functions that are invariant to involutory transformations. In Section \ref{sec:associated}, we present associated techniques based that can be used to handle multiple symmetries and show how to construct invariant kernels for CNNs. Section \ref{sec:exp} contains experimental results on an array of test problems, and a discussion ensues in Section \ref{sec:discuss}. Future directions of research are discussed in Section \ref{sec:future} and we conclude in Section \ref{sec:conclusion}

\section{Hub-Layered Network (HLN)}
\label{sec:hub.recap}

\cite{mattheakis.symmetry} define a hub-layered NN  that respects an even/odd symmetry about a fixed point and learns a function $f:\mathbb{R}\rightarrow\mathbb{R}$. An even function in 1-D is invariant to the involutory transformation $-I_1=[-1]_{1\times1}$ with parity $p=1$. Similarly, an odd function is invariant to $-I_1$ but with $p=-1$.

Let the activation value of $i^{\text{th}}$ node in the final hidden layer of the NN be $h_i$. It represents some function of the input $x$ presented to the neural network, denoted as $h_i(x)$. The output $\mathit{NN}(\cdot)$ of a vanilla neural network would be
\begin{align}
    \mathit{NN}(x) = \sum_{i}w_ih_i(x) + b. \label{eq:vanilla.output}
\end{align}
Define a hub-layer as $H^{\pm}(x)=h_i(x)\pm h_i(-x)$. A neural network $H_{\text{even}}(\cdot)$ that learns even functions could be constructed as follows
\begin{align}
    H_{\text{even}}(x) = \sum_i w_i H^+(x) + 2b. \label{eq:hub.even.output}
\end{align}
The name \emph{hub layer} is intended to indicate the fact that the final layer $H^+$ is a superposition of $h_i(x)$ and $h_i(-x)$ of an otherwise vanilla neural network. Eq. (\ref{eq:hub.even.output}) can be shown to satisfy $H_{\text{even}}(-x) = H_{\text{even}}(x)$. An odd function can be analogously represented as
\begin{align}
    H_{\text{odd}}(x) = \sum_iw_iH^-(x). \label{eq:hub.odd.output}
\end{align}
The bias $b$ has been removed due to the requirement that the function learnt be odd. Eq. (\ref{eq:hub.odd.output}) can be shown to satisfy $H_{\text{odd}}(-x)=-H_{\text{odd}}(x)$. The architecture is easily extended so that it can learn functions $\reals{n}\rightarrow\mathbb{R}$ of $n-$dimensional input, that are invariant to an involutory transformation, $A \in \reals{n\times n}$, up to parity $p=\pm1$. More specifically,  the network must satisfy the symmetry requirement
\begin{align}
    \mathit{NN}(X) = p\mathit{NN}(AX). \label{eq:involutory.symm.requirement}
\end{align}
If one formulates the network output as
\begin{align}
    H^p(X) &= h_i(X) + ph_i(AX), \label{eq:hub.involutory.layer} \\
    \mathit{NN}_{A;p}(X) &= \sum_iw_iH^p(X) + \theta(p)\times2b, \label{eq:hub.involutory.output}
\end{align}
the invariance $\mathit{NN}_{A;p}(AX)=p\mathit{NN}_{A;p}(X)$ is met. Here, $\theta(\cdot)$ represents the Heaviside step function and acts on the parity $p=\pm1$

\section{Symmetrized Activation Network (SAN)}
\label{sec:san}

We consider only $p=1$ for this section. From Eq. (\ref{eq:hub.involutory.layer}), the HLN  requires two forward passes up to the last hidden layer \textit{i.e.}, $h_i(X)$ and $h_i(AX)$,  before they can be summed by the inference layer as in Eq. (\ref{eq:hub.involutory.output}). This takes $2\times$ more computational time compared to a standard NN. Hence, the guarantee of symmetry comes at an overhead of one extra forward pass through the network. This overhead grows proportionally to the depth $d$ of the network as $\Theta(d)$. We can reduce the overhead by the following observation

\textit{To guarantee that the network output is invariant to an involutory transformation, it is sufficient to ensure that \textbf{all} the activations in \textbf{some} hidden layer are invariant to the involutory transformation}

For minimum overhead, we allow double forward passes only upto the first hidden layer. This reduces the overhead to $\Theta(1)$ as it is independent of the network depth. In other words, we require $a^{[1]}_i(X)=a^{[1]}_i(AX)$ where $a^{[1]}_i$ represents the activation of the $i^{\text{th}}$ node in the first hidden layer. 

\textbf{Example:} In the case of even functions on the real number line $\mathbb{R}$, this requires $a^{[1]}_i(-x)= a^{[1]}_i(x)$. It is possible to achieve this by superimposing a linear sum of the activation function $\sigma(\cdot)$ as
\begin{align}
    a^{[1]}_i(x)=\sigma(w_ix+b) + \sigma(-w_ix+b),
\end{align}
and it can be seen that $a^{[1]}_i(-x)= a^{[1]}_i(x)$ holds. Extending this to generic $n$-dimensional case, one can use
\begin{align}
    a^{[1]}_i(X)=\sigma(w^TX+b) + \sigma(w^TAX+b).
\end{align}
It is easily seen that $a^{[1]}_i(AX)=a^{[1]}_i(X)$. Such a superposition of activation functions as the one given above is called a \emph{symmetrized activation} function.

\subsection{The No-Expressivity Problem}
\label{sec:safe}

In the initial phase of experimentation with SAN, 
the bias parameter of the first layer was fixed to zero and $\texttt{sigmoid}$ activation was used. It was noted that the inference output was independent of the input point $x$ --- there was a lack of expressivity in the network's output.

Let the activation value of the first layer be $a^{[1]}_i(x) = \sigma(w_ix) + \sigma(-w_ix)$. Consider the variable $z_i=w_ix$ for some weight $w_i$. Since $a^{[1]}_i(x)$ is even, the network output is also even. It is a property of $\texttt{sigmoid}$ that
\begin{align}
    \sigma(z) + \sigma(-z) = \frac{1}{1+e^{-z}} + \frac{1}{1+e^z} &= 1 \label{eq:sigmoid.fact1} \\
    \implies \dv{}{z}\left[\sigma(z)+\sigma(-z)\right] &= 0 \label{eq:sigmoid.fact2}
\end{align} due to which
\begin{align}
    \pdv{a_i^{[1]}}{x} = \pdv{a_i^{[1]}}{w_i} = 0 \label{eq:sigmoid.frozen}
\end{align}
Hence the weight $w_i$ does not change during training. Moreover, since $i$ was chosen arbitrarily, all the weights in the first layer do not change. From Eq. (\ref{eq:sigmoid.frozen}), the NN learns a constant function because all activations in the first layer are independent of the input $x$. This is precisely the lack of expressivity that was encountered in our initial experimentation.

This situation can be generalized to the $n$ dimensional case. For $X,w_i\in\reals{n}$, and $b\in\mathbb{R}$, the activation value of the $i^{\text{th}}$ node in the first hidden layer
\begin{align}
    a_i^{[1]}&=\sigma(b+w_i^TX) + \sigma(b+w_i^TAX), \label{eq:act.ith}
    \\
    \nabla_X a_i^{[1]} &= [\sigma'(b+w_i^TX)I_n \nonumber \\
    &+ \sigma'(b+w_i^TAX)A^T]w_i, \label{eq:act.ith.der}
\end{align}
where by $\sigma'$ we mean the derivative of $\sigma$. Let $z=w_i^TX$ be a uni-dimensional variable. It covers the full range of $\mathbb{R}$ as we have premised that $X, w_i \in \mathbb{R}$ arbitrary. For the NN to retain expressivity, we need $\nabla_X a_i^{[1]}\neq\va{0}_n$. 

For the sake of argument, let assume that expressivity is lost, \textit{i.e.} $\nabla_X a_i^{[1]}=\va{0}_n$. This is only possible if the matrix within the braces $[ \ ]$ in Eq. (\ref{eq:act.ith.der}) is the zero matrix. The $\sigma'(\cdot)$ terms are scalar multipliers to the matrices $I_n$ and $A^T$. Thus, the matrix $A$ cannot have any off-diagonal non-zero entries. Moreover, the only scenario in which our assumption can be satisfied
\begin{align}
A=-I_n. \label{eq:no.expressivity.cond}
\end{align}
We can now expand on Eq. (\ref{eq:act.ith.der}) as follows
\begin{align}
   \sigma'(b+w_i^TX) - \sigma'(b-w_i^TX) &= 0 \label{eq:sigma.der.zero} \\
   \int \sigma'(b+z)\dd{z} - \int \sigma'(b-z)\dd{z} &= 0 \nonumber \\
   \sigma(b+z) + \sigma(b-z) &= C, \label{eq:integration.constant}
\end{align}
where $C\in\mathbb{R}$ is a constant of integration. Since $z$ is a free variable, the only way Eq. (\ref{eq:integration.constant}) will be satisfied is if the bias is assumed to take some special value $b^*$, in which case we could state
\begin{gather}
      \left[\sigma(b^*+z) - \frac{C}{2}\right] = -\left[ \sigma(b^*-z)-\frac{C}{2}\right]. \label{eq:odd.derivation}
\end{gather}
Substituting $z=0$ (\emph{since $z$ is a free variable}), we get $C=2\sigma(b^*)$. If we define a new function $\sigma_{b^*}(z) = \sigma(b^*+z) - \sigma(b^*)$, we can rewrite Eq. (\ref{eq:odd.derivation}) as
\begin{align}
    \sigma_{b^*}(-z) = -\sigma_{b^*}(z). \label{eq:odd.star.condition}
\end{align}
The interpretation of Eq. (\ref{eq:odd.star.condition}) is as follows --- \emph{The activation function $\sigma$ is odd about some $b^*$ after a shift by $\sigma(b^*)$}

During training, if the bias assumes the value of $b^*$, the training freezes because of Eq. (\ref{eq:no.expressivity.cond}) and Eq. (\ref{eq:sigma.der.zero})
\begin{align}
    \nabla_{w_i} a_i^{[1]} &= \left[\sigma'(b+w_i^TX)I_n + \sigma'(b+w_i^TAX)A^T\right]X \nonumber \\
    &= [\sigma'(b+w_i^TX) - \sigma'(b-w_i^TX)]X \label{eq:midstep} \\
    &= \va{0}_n \label{eq:ndimensional.freeze}
\end{align}
for all nodes $i$ in the first hidden layer. In Eq. (\ref{eq:midstep}), we have assumed $A=-I_n$ as was argued previously. This limits the NN to only learn constant functions from $\reals{n}\rightarrow\mathbb{R}$.

One needs to worry about the \emph{no-expressivity} problem only when the involutory invariance being learnt is an inversion, and the parity $p=1$. Moreover, it explains why fixing the bias to zero leads to a lack of expressivity as $b=0$ happens to be one of the special (\emph{in fact, the only}) values $b^*$ for \texttt{sigmoid} activation
\begin{gather}
    \eval{\sigma_{b^*}(z)}_{b^*=0} = \frac{1}{1+e^{-z}} - 1 = \eval{\sigma_{b^*}(-z)}_{b^*=0} \label{eq:1d.noexpress.verdict}
\end{gather}

\subsection{Symmetrized Activation Classes}

We define a class of functions, which if used as a symmetrized activation function for training, can cause a lack of expressivity when learning an inversion invariance with even parity. We obtain the following definition by obeying Eq. (\ref{eq:odd.star.condition})
\begin{align*}
    \texttt{INV\mbox{-}UNSAFE$^+$} &= \{
    \sigma \mid \sigma : \mathbb{R} \rightarrow \mathbb{R} ; \exists \ b^* : \sigma_{b^*}(z) \\ &\text{is odd where} \ \sigma_{b^*}(z) = \sigma(b^*+z) - \sigma(b^*)\} \\
    \texttt{INV\mbox{-}SAFE$^+$} &= \{
    \sigma \mid \sigma : \mathbb{R} \rightarrow \mathbb{R} ; \nexists \ b^* : \sigma_{b^*}(z) \\ &\text{is odd where} \ \sigma_{b^*}(z) = \sigma(b^*+z) - \sigma(b^*) \}
\end{align*}
By definition, the two classes are complements of each other. We have $\texttt{sigmoid}\in\texttt{INV\mbox{-}UNSAFE$^+$}$. The $+$ in the superscript refers to even parity $p=1$

\textbf{Example:} In \cite{snake}, a variant of the $\texttt{snake}$ activation function is used where $\sigma(z)=z+\sin z$ which has infinitely many \emph{unsafe} points $b^*=\pm n\pi$ for $n\in\mathbb{W}$
\begin{align}
    \sigma_{b^*}(z) &= \sigma(\pm n\pi+z) - \sigma(\pm n\pi) \nonumber \\
    &= (\cancel{\pm n\pi} + z) + \sin(\pm n\pi + z) \cancel{\mp n\pi} - \sin(\pm n\pi) \nonumber \\
    &= (-1)^{n}\sin(z) \nonumber \\
    &= -\sigma_{b^*}(-z) \label{eq:snake.odd}.
\end{align}
Thus, $\texttt{snake} \in \texttt{INV\mbox{-}UNSAFE$^+$}$ because of Eq. (\ref{eq:snake.odd}). Similarly, $\texttt{tanh} \in \texttt{INV\mbox{-}UNSAFE$^+$}$ as it is odd about the origin.

\textit{\textbf{Result:}} $\texttt{ReLU}$ \cite{relu}, $\texttt{swish}$ \cite{swish}, $\texttt{softplus}$ \cite{softplus} $\in \texttt{INV\mbox{-}SAFE$^+$}$.

\textit{\textbf{Proof:}} Assume $\exists b^*$ such that $\sigma_{b^*}(z) = \sigma(b^*+z)-\sigma(b^*)$ is odd. The concerned activation functions, \texttt{ReLU}, \texttt{swish} and \texttt{softplus} have the following behaviour
\begin{align}
    \lim_{z\rightarrow \infty}\sigma_{b^*}(z)&=\infty \\
    \lim_{z\rightarrow -\infty}\sigma_{b^*}(z)&=0
\end{align}
Due to their asymptotic/diverging behaviour as $z\rightarrow \pm\infty$ , it is possible to state (\emph{with an $L$ sufficiently large and $\epsilon$ sufficiently small})
\begin{align}
    \exists \ L>1, \exists \ z_+>0 : z \geq z_+ &\implies \sigma_{b^*}(z) \geq L \\
    \exists \ 0\leq\epsilon<1, \exists \ z_-<0 : z \leq z_- &\implies |\sigma_{b^*}(z)| \leq \epsilon
\end{align}
Define a new variable $z^*=max(|z_+|, |z_-|) \implies z^*\geq z_+$ and $-z^*\leq z_-$. This would imply
\begin{gather}
    1 < L \leq \sigma_{b^*}(z^*) \ \textit{and} \  |\sigma_{b^*}(-z^*)| \leq \epsilon < 1 \\
    \implies \sigma_{b^*}(-z^*) \neq  -\sigma_{b^*}(z^*)
\end{gather}
Hence, for any $b^*$ it is always possible to find a point $z^*$ about which the function fails to be odd. This contradicts the fact that $\sigma_{b^*}(\cdot)$ was odd, and hence our assumption on the existence of a $b^*$ was incorrect. Thus 
\texttt{ReLU, swish, softplus} $\in \texttt{INV\mbox{-}SAFE$^+$}$.

A similar analysis can be performed for functions obeying inversion invariance with odd parity $p=-1$. Analogously, we define two additional classes of activation functions
\begin{align*}
    \texttt{INV\mbox{-}UNSAFE$^-$} &= \{
    \sigma \mid \sigma : \mathbb{R} \rightarrow \mathbb{R} ; \exists \ b^* : \sigma_{b^*}(z) \\ &\text{is even where} \ \sigma_{b^*}(z) = \sigma(b^*+z)\} \\
    \texttt{INV\mbox{-}SAFE$^-$} &= \{
    \sigma \mid \sigma : \mathbb{R} \rightarrow \mathbb{R} ; \nexists \ b^* : \sigma_{b^*}(z) \\ &\text{is even where} \ \sigma_{b^*}(z) = \sigma(b^*+z)\}
\end{align*}

\textbf{Subtlety:} It is not necessarily true that using an activation function $\sigma\in\tt{INV\mbox{-}UNSAFE^{\pm}}$ would \emph{always} cause a lack of expressivity. In practice, we have found that $\tt{sigmoid}$ is capable of learning functions that obey an inversion invariance with even parity if the bias is initialized to a value far from $b=0$. This is distinctly different from $\tt{snake}$, however, as it has infinitely many \emph{unsafe} points, and hence a \emph{poor} choice for a symmetrized activation function.

\section{IPT Network (IPTNet)}
\label{sec:iptnet}

To design an efficient involutory-invariant NN architecture, it is important to avoid double-computations \emph{altogether} as compared to the SAN Network. The motivation behind the involutory partition theorem can be found by considering the instance of learning a 1D function. 
If the function is even/odd, it is only required to specify its behaviour in $\mathbb{R}_+ \cup \{0\}$. If there happens to be a training set example $(x^{(i)}, y^{(i)})$ where $x^{(i)} < 0$, it can be fed to the NN as $(-x^{(i)}, p\times y^{(i)})$ instead. The entire training set can be modified in this fashion prior to training. During inference, if we require the NN output on a point $z$, we \emph{reparameterize}\footnote{for an introduction to the technique of reparameterization, refer to \cite{lagaris}}
it as follows
\begin{align}
    \mathit{NN}(z;p) \leftarrow \begin{cases}
    \mathit{NN}(z) & z \geq 0 \\
    p\times \mathit{NN}(-z) & z < 0
    \end{cases} \label{eq:ipt.reparam.1d}
\end{align}
Any negative input is flipped in sign and then fed for inference. The reparameterization \emph{encodes} the fact that network's output is solely determined by its behaviour in $\mathbb{R}_+ \cup \{0\}$, and in turn \emph{encodes} the fact that the function learnt is even/odd.

Since even/odd functions are linked to $-[1]_{1\times 1}$, one could say that the matrix $-I_1$ has partitioned the real number line in three mutually exclusive sets $(\mathbb{R}_-, \{0\}, \mathbb{R}_+)$. It would be fruitful if we could find an analogous partitioning of $\reals{n}$ into subsets $(S_-, S_0, S_+)$ induced by an arbitary involutory transformation $A\in\reals{n\times n}$. If such a partitioning can be found, we can \emph{encode} all involutory invariances by an appropriate reparameterization
\begin{align}
    \mathit{NN}(Z;p) \leftarrow \begin{cases}
    \mathit{NN}(Z) & Z \in S_0 \cup S_+ \\
    p\times \mathit{NN}(AZ) & Z \in S_-
    \end{cases} \label{eq:ipt.reparam.ND}
\end{align}
The $\mathit{NN}$ written above could be any standard feed-forward neural network. In other words, the IPTNet is essentially an FFNN encapsulated with a reparameterization of its input.

\subsection{Involutory Partition Theorem (IPT)}
\textit{\textbf{Theorem:} An involutory matrix $A \neq I_n$ partitions $\reals{n}$ into three mutually exclusive non-empty subsets $S_0, S_+, S_-$, with the following properties:}
\begin{align*}
    &1. \myforall \va{v} \in S_0, A\va{v} = \va{v} \\
    &2. \myforall \va{v} \in S_+, A\va{v} \in S_- \\
    &3. \myforall \va{v} \in S_-, A\va{v} \in S_+
\end{align*}

\textit{\textbf{Proof Sketch:}} Although the proof of the IPT relies on simple mathematics, we have avoided stating the full formal proof for brevity. Instead, we provide a proof sketch and an intuition for its correctness. The full proof can be provided on request from the authors.

\noindent\textit{Case 1} --- $A=-I_n$

We provide a construction for the partitioning sets $(S_-, S_0, S_+)$. For statement 1, the only vector that satisfies this condition is the zero vector $\va{0}_n$. Next, we show a recursive construction of $S_+$ and $S_-$. Consider the final $n^{\text{th}}$ dimension only. Let $S_+$ consist of all vectors $\va{v}\in\reals{n}$ with the final dimension $>0$ (\emph{strictly positive}). Hence, for statement 2/3 to remain valid $S_-$ is bound to consist of all vectors with the final dimension $<0$ (\emph{strictly negative}). We are now left with all vectors $\va{v}$ with the final dimension as $0$. This forms an $(n-1)$ dimensional subspace and the involutory transformation that acts on this subspace is $-I_{n-1}$. We can repeat a similar argument, assort vectors into $S_+$ or $S_-$ considering the first $(n-1)$ dimensions only, and then recurse until the single dimensional case is reached. 

\noindent\textit{Case 2} --- $A\neq-I_n$

We exploit a standard result from matrix theory --- All involutory matrices are diagonalizable to the form $D=diag(1, 1, \ldots, -1, -1, \ldots, -1)$. The derivation of this result can be found in \cite{matrix.theory}.

Let the number of $-1$'s in this diagonal form be $\gamma$. On the first $(n-\gamma)$ dimensions, the transformation acts as an identity $I_{n-\gamma}$ in the appropriate $\reals{n-\gamma}$ subspace. All vectors strictly belonging to this subspace are part of the set $S_0$. The remaining vectors have non-zero components in the last $\gamma$ dimensions and the action of $D$ in this subspace is that of an inversion transformation $-I_\gamma$. Thus, they can be partitioned into $S_+$ or $S_-$ appropriately using the aforementioned recursive construction. 

If the diagonalizing matrix is $P$, we have $D=P^{-1}AP$. However, diagonalization acts as a linear transformation in the space of vectors and has no effect on the partitioning $(S_-, S_0, S_+)$. Hence, it is sufficient to prove the IPT for the diagonal form $D$ as it covers the case of all involutory transformations $A\neq-I_n$. This completes the proof sketch, and it can be confirmed that the network defined by Eq. (\ref{eq:ipt.reparam.ND}) satisfies the invariance requirement Eq. (\ref{eq:involutory.symm.requirement}). 

\subsection{Membership Algorithm}

For an involutory matrix $A$, we define its \emph{Principal Involutory Domain}, $PID(A)=S_0 \cup S_+$. We now require an efficient algorithm that determines whether an input vector belongs to the $PID$ so that the reparameterization (Eq. \ref{eq:ipt.reparam.ND}) can be implemented. 

\begin{codebox}
\Procname{\proc{VECTOR-IN-PID}($\va{v}, \id{Pinv}, \gamma, n$)}
\li \For $\va{r}$ in $\id{Pinv.rows}[n, \ldots, n-\gamma+1]$
    \li \Do $\id{e} \gets \va{r} \cdot \va{v}$
    \li \If $\id{e}>0$
        \li \Then \Return $\id{True}$
    \li \ElseIf $\id{e}<0$
        \li \Then \Return $\id{False}$
    \li \ElseNoIf \textbf{continue}
    \End
\End
\li \Return $\id{True}$
\label{algo:pid.membership}
\end{codebox}

\textbf{Description:} $\va{v}$ is the input vector whose membership is to be determined. The argument $\id{Pinv}$ is the matrix $P^{-1}$, which is the inverse of the diagonalizing matrix for $A$. $\gamma$ is the number of $-1$ entries in the diagonal form $D$ of $A$. Finally, $n$ is the number of dimensions of vector $\va{v}$. The for-loop iterates over the rows of $P^{-1}$ in reverse order and computes a dot-product $\va{r}\cdot\va{v}$ which is subject to an if-else condition.

\textbf{Time Complexity:} Computing the dot product takes $\Theta(n)$ time, and there are at-most $\gamma$ such dot products to be computed. This gives the algorithm an asymptotically tight run-time of $\Theta(n\gamma)$. Since $n \geq \gamma=\bigoh{n}$, it results in an upper bound of $\bigoh{n^2}$. This cost need not be incurred every training iteration. The reparameterization can be pre-computed by relocating training points to the $PID$ if required. This could be performed for each point in the training set, and the modified training set be used instead for learning. This is a constant fixed cost, and hence can be ignored. In the case of obtaining inference at new test points, however, the $\Theta(n\gamma)$ cost has to be incurred. 

\textbf{Correctness:} In a nutshell, the procedure transforms an input vector to the diagonalized coordinates (\emph{by multiplying as $P^{-1}\va{v}$}) and checks for the first positive/negative entry in the new vector in reverse order. Accordingly, the vector is adjuged to be in $S_+, S_-$. If the for-loop completes, then the vector belongs to $S_0$ in which case $\id{True}$ is returned.

\subsection{Involutory Invariant Universal Approximation Theorem}

\textit{\textbf{Theorem:} With the appropriate reparameterization, single hidden-layer neural networks can approximate to arbitrary precision, any function $f:\reals{n}\rightarrow\mathbb{R}$ that respects an involutory invariance $f(AX)=pf(X)$ for some $A\in\reals{n\times n}$ and parity $p=\pm 1$ with $A\neq I_n$}.

\textit{\textbf{Proof:}} It is best encapsulated in the following logical series of statements
\begin{enumerate}
    \item A function with an involutory invariance is completely specified by its behaviour in its $\pid$.
    \item All training examples can be transformed into the $\pid$ due to the reparameterization, and the $\pid$ is a domain in the mathematical sense.
    \item FFNNs can approximate, to arbitrary precision, any function in a given domain provided all training examples belong to it \cite{cybenko}.
    \item Since the IPTNet is essentially an encapsulated FFNN it can approximate, to arbitrary precision, any involutory invariant function in $\reals{n}$.
\end{enumerate}
\textbf{Subtlety:} One might ask how the involutory UAT is any \emph{better} than the standard UAT. The general UAT statement holds in our setting because involutory invariant functions are a subset of the general class of functions that single-layered NNs can learn. Although, any function can be learnt to arbitrary precision by a standard FFNN, there is no guarantee that invariance requirement is met simultaneously. The involutory UAT guarantess both arbitrary precision and invariance. In a nutshell, the key intuition is

\emph{Standard UAT in $PID\implies $Involutory Invariant UAT}

\section{Extensions of the IPT}
\label{sec:associated}

\subsection{Symmetry to Involutory Affine Transformations}
\label{sec:affine}

It is possible to learn functions invariant to involutory \emph{affine} transformations as well. A transformation $T$ is said to be affine if it is of the form $T(X)=AX+\mu$ for an input $X\in\reals{n}$, and $A\in\reals{n\times n}, \mu\in\reals{n}$. They satisfy $A^2=I$ and $A\mu = -\mu$. This property can be exploited to perform a shift of the origin as $X'\gets X-\mu/2$ and $T'\gets T-\mu/2$ which results in
\begin{align}
    T'(X') = AX' \label{eq:affine.to.linear}
\end{align}
Thus, the action of an involutory \emph{affine} transformation has been converted to that of an involutory \emph{linear} transformation after a shift of the origin by $\frac{\mu}{2}$. The theorems proved, $PID$ membership algorithm and the IPT-Net can now be used to the same effect.

\subsection{Handling Multiple Involutory Invariances}
\label{sec:multiple}

Consider a function that respects reflection about both the X and Y axes independently, such as function $f(x,y)=x^2+\cos{y}$. An HLN that is invariant with respect to both is constructed as
\begin{align}
    H(x, y) &= h(x,y) + h(x,-y) \nonumber \\
    & + h(-x,y) + h(-x,-y). \label{eq:superimpose4}
\end{align}
Note that $H(x,y)$ is \emph{independently} invariant to both $x\rightarrow-x$, and $y\rightarrow-y$. However, this incurs a $4\times$ overhead up to the final hidden layer to guarantee symmetry. 

\textbf{Example:} In 3D, a function could respect inversion in $(x,y)$ with odd parity and be even in $z$ \emph{independently}. For instance, $f(x,y,z)=z^2(x+y)$ follows the invariances
\begin{align*}
f(-x,-y,z) &= -f(x,y,z), \\
f(x,y,-z) &= f(x,y,z).    
\end{align*}
There are two involutory matrices at play. $A_1$ is acting on the $(x,y)$ dimensions with parity $p_1=-1$, while  $A_2$ is acting on the $z$ dimension with $p_2=1$, hence
\begin{align*}
    A_1 = \begin{bmatrix}
    -1 & 0 \\
    0 & 1
    \end{bmatrix}_{(x,y)}, \quad A_2 = [-1]_{z}.
\end{align*}
\textbf{Generalization:} Let $A_1\in\reals{n_1\times n_1}$,  $A_2\in\reals{n_2\times n_2}, \ldots,$ $A_k\in\reals{n_k\times n_k}$ be the list of involutory transformations that a function $f : \reals{n} \rightarrow \mathbb{R}$ has to respect $\left(\sum_{i=1}^kn_i=n\right)$ with parities $p_1, p_2, \ldots, p_k$. Each of the $A_i$ acts upon a subset of dimensions mutually exclusive from any other $A_j$. To analyse the upper bound on multiple hidden layer computations required by the hub layer, we consider the case where each dimension \emph{independently} respects a sign flipping invariance ($k=n$). This implies all $n_i=1$ and one would incur a $2^n\times$ extra cost\footnote{
There would be $2^n$ terms in the expression for the hub-layer because each dimension $x_i$ can independently appear as $\pm x_i$ in $H(\cdot, \cdot, \ldots, \cdot)$
} 
on the hidden layer computations. Hence, hub-layered networks have an exponential $\bigoh{2^n}$ cost of guaranteeing invariance in the worst case. It is possible to reduce the naive $\bigoh{2^n}$ hidden layer cost to an $\Theta(k)$ cost by an altered architecture using the IPT. The idea is to use the cross-product of the principal involutory domains of each $A_i$ as 
\begin{align}
\otimes_{i=1}^k \pid(A_i)
\end{align}
Since there are $k$ such transformations, the cost of guaranteeing invariance is $\Theta(k)$.

\subsection{Reflection Invariant Kernels}

For certain CNN tasks, if it is known beforehand that the images obey an X-reflection or Y-reflection symmetry, it is possible to create reflection invariant feature maps by the superposition of filters akin to the technique of symmetrized activation functions. An immediate use-case of such a kernel would be exploiting the vertical symmetry in human face datasets. The technique is best in code
$$
\sigma(\texttt{filter} * \texttt{image}) + \sigma(\texttt{filter} * \texttt{image.flip(axis=1))} \label{eq:pytorch.code}
$$
where $*$ is the convolution operation, and $\sigma$ is an optional non-linearity. If the first layer of convolutions are implemented in this way, it can be easily seen that the feature map computed is invariant to the operation $\texttt{image.flip(axis=0)}$. To ensure invariance to X-reflections of the input image, one would use \texttt{flip(axis=1)} instead. \footnote{Note that the presence of a pooling layer, which is standard in CNNs, does not affect the overall invariance}

\section{Experiments}
\label{sec:exp}

\subsection{Toy Functions}
\label{sec:toy}

For this subsection and the next, we test the following architectures
\begin{enumerate}
    \item Vanilla Network (\emph{VN})
    \item Hub Layered Network (\emph{HLN})
    \item Symmetrized Activation Network (\emph{SAN}) with a symmetrized \texttt{swish} activation in the first layer
    \item IPTNet (\textbf{IPTN})
\end{enumerate}

Consider the task of learning the function $\cos x$ where 100 points are uniformly sampled from the interval $[-1.5, 1.5]$. The target value $\cos(\cdot)$ for each training point is varied by a weak gaussian noise. The four networks all use 2 hidden layers with 10 neurons each, and are trained for 5000 epochs with a learning rate of $0.005$, MSE loss and \texttt{Adam} optimizer. The activations used in the hidden layers is \texttt{sigmoid}, unless specified otherwise.

We construct a validation set of uniformly distributed points in $[-3, 3]$ and devise a \emph{violation metric} on this set as follows
\begin{gather}
    V_p = \frac{1}{m} \sum_{i=1}^m (\mathit{NN}(x) - p\mathit{NN}(-x))^2,
\end{gather}
where $p=\pm 1$. For $\cos x$, we have $p=1$ and $V_+$ measures how much the predicted by $\mathit{NN}$ function deviates from being an even function.

\begin{figure}[h]
    \centering
    \caption{Training curves of $\cos x$ }
    \label{fig:cosx}
    \begin{subfigure}{.25\textwidth}
        \centering
        \includegraphics[width=.9\linewidth]{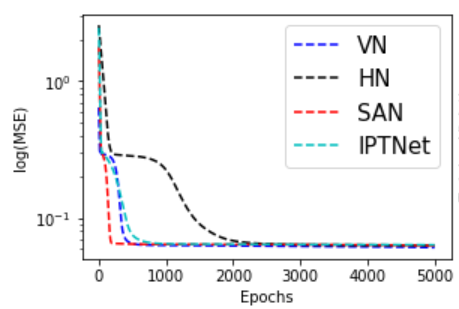}
        \caption{log(MSE) vs. epochs}
        \label{fig:1d.logmse}
    \end{subfigure}%
    \begin{subfigure}{.25\textwidth}
        \includegraphics[width=.9\linewidth]{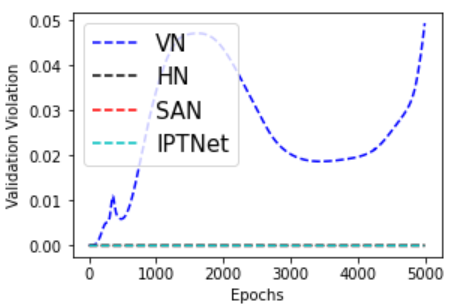}
        \caption{Violation metric vs. epoch}
        \label{fig:1d.valvio}
    \end{subfigure}
\end{figure}

\begin{figure}[h]
    \centering
    \caption{Contours of $\sin x + \sin y$}
    \label{fig:2d.contours}
    \begin{subfigure}{.25\textwidth}
        \centering
        \includegraphics[width=.8\linewidth,height=2.5cm]{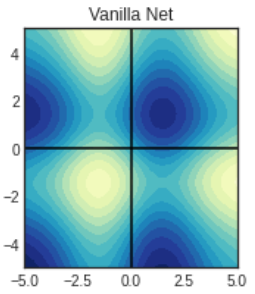}
        \caption{VN Contour}
        \label{fig:contour.vn}
    \end{subfigure}%
    \begin{subfigure}{.25\textwidth}
        \includegraphics[width=.8\linewidth,height=2.5cm]{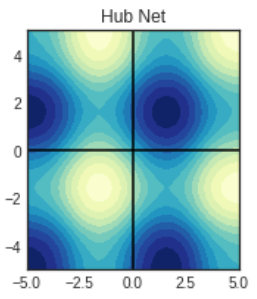}
        \caption{HLN contour}
        \label{fig:contour.hln}
    \end{subfigure}
    \begin{subfigure}{.25\textwidth}
        \hspace{0.21in}\includegraphics[width=.8\linewidth,height=2.5cm]{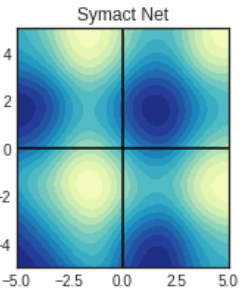}
        \caption{SAN contour}
        \label{fig:contour.san}
    \end{subfigure}%
    \begin{subfigure}{.25\textwidth}
        \includegraphics[width=.80\linewidth,height=2.5cm]{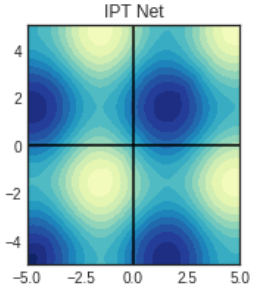}
        \caption{IPT contour}
        \label{fig:contour.ipt}
    \end{subfigure}
\end{figure}

Table \ref{table:1d.loss.vio} shows the mean loss and violation metrics over 25 independent instance of training, along with the respective standard deviations, where the number in the subscript represents the standard deviation. Fig. \ref{fig:cosx} tracks the MSE and violation metric over the epochs of training.
The invariant networks (HLN, SAN, IPTN) obey the symmetry exactly as shown by a violation metric of $0$. VN, on the other hand, shows large deviations from symmetry. From table \ref{table:1d.fwd.back}, the IPTNet records the lowest times (\emph{barring VN}) in forward-propagation and backprop since it has the least overhead of encoding symmetry. HLN and SAN require more time as the former includes double computations upto the final layer, and the latter only has double computation upto the first layer. Fig. \ref{fig:2d.contours} shows the contours of a 2D function learnt by the proposed networks.
\begin{table}[htb]
\centering
\caption{Loss and Violation}
\label{table:1d.loss.vio}
\begin{tabular}{|c|c|c|}
\hline
    & Loss (MSE)                                       & Violation                                  \\ \hline
VN  & $6.14\times 10^{-2}_{ 3.18\times10^{-8}}$  & $3.67\times 10^{-2}_{ 1.19\times 10^{-2}}$ \\ \hline
HLN & $9.45\times 10^{-2}_{ 7.42\times 10^{-2}}$ & $0.0_{ 0.0}$                               \\ \hline
SAN & $6.29\times 10^{-2}_{ 5.55\times 10^{-4}}$ & $0.0_{ 0.0}$                               \\ \hline
\textbf{IPTN} & $6.27\times 10^{-2}_{ 1.94\times 10^{-4}}$ & $0.0_{ 0.0}$           \\ \hline
\end{tabular}
\end{table}
\begin{table}[htb]
\centering
\caption{Forward/Back Propagation Times (in seconds)}
\label{table:1d.fwd.back}
\begin{tabular}{|c|c|c|}
\hline
    & Avg FP               & Avg BP               \\ \hline
VN  & $8.44\times 10^{-5}$ & $1.37\times 10^{-4}$ \\ \hline
HLN & $1.77\times 10^{-4}$ & $2.42\times 10^{-4}$ \\ \hline
SAN & $1.95\times 10^{-4}$ & $1.94\times 10^{-4}$ \\ \hline
\textbf{IPTN} & $1.35\times 10^{-4}$ & $1.50\times 10^{-4}$ \\ \hline
\end{tabular}
\end{table}

\subsection{Learning Physical Dynamics}

We build upon the Hamiltonian Neural Network (HNN) model by \cite{hnn} and run experiments on a physics-inspired setting, the ideal spring, which is represented by the Hamiltonian 
\begin{gather}
    H(q,p) = \frac{1}{2}kq^2 + \frac{p^2}{2m}    
\end{gather}
The input dataset is a set of noisy observations $(q, p)$ and $(\dot{q}, \dot{p})$ from the phase space, from which the  NN constructs a quantity akin to the Hamiltonian $H_{\theta}$. With the help of automatic differentiation, the network optimizes the following loss function
\begin{gather}
L_{\textit{HNN}} = \left(\pdv{H_{\theta}}{p} - \dot{q}\right)^2 + \left(\pdv{H_{\theta}}{q} + \dot{p}\right)^2
\end{gather}
$H(q,p)$ is invariant with respect to both $q \rightarrow -q$ and $p \rightarrow -p$. Thus, the tranining set can be modified as per the prescription of Section \ref{sec:multiple} (\emph{multiple symmetries involved}) and the sign invariance of $(q, p)$ can be encoded in the HNN.
\begin{figure}[h]
    \centering
    \caption{Ideal Spring}
    \label{fig:hnn}
    \begin{subfigure}{.25\textwidth}
        \centering
        \includegraphics[width=.8\linewidth]{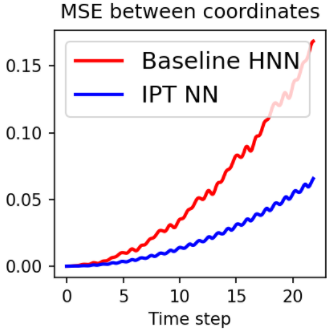}
        \caption{Coordinates MSE}
        \label{fig:hnn.mse}
    \end{subfigure}%
    \begin{subfigure}{.25\textwidth}
        \includegraphics[width=.8\linewidth]{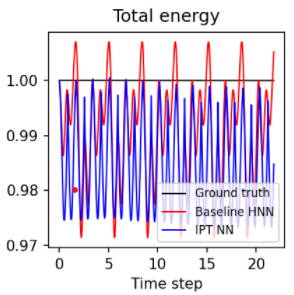}
        \caption{Energy curve}
        \label{fig:hnn.energy}
    \end{subfigure}
\end{figure}
The results are shown in Fig. (\ref{fig:hnn}). Encoding the sign inversion shows significant reduction in the MSE deviation of coordinates. Moreover, the predicted energy of the IPT-HNN model oscillates less, namely the error variance is smaller than the baseline HNN. Thus, encoding the sign invariance aspect conserves energy better.

\subsection{Image Tasks}
\label{sec:fashion.mnist}

\begin{figure}[h]
    \centering
    \caption{Samples from the Yale face dataset}
    \includegraphics[width=0.25\textwidth]{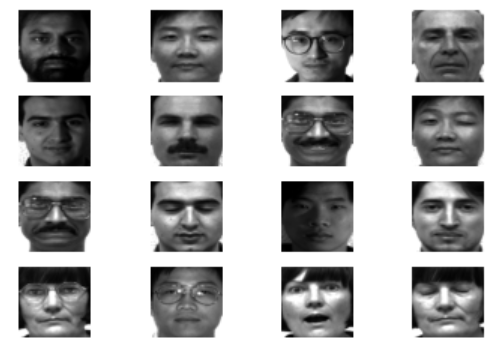}
    \label{fig:yale.face}
\end{figure}

For CNN tasks, we chose a dataset that has an inherent horizontal/vertical symmetry, such that the symmetry is consistently represented across all training/validation examples. The Yale face dataset \cite{yale.face} has 165 grayscale images of 15 individuals. The faces exhibit a natural vertical symmetry as shown in Fig. (\ref{fig:yale.face}). 
We compare two NNs one of which is a vanilla CNN (VCNN) that serves as the baseline. The second model is identical to the baseline except that the first layer of convolutions is fitted with Y-reflection invariant kernels. It is called the invariant-kernel CNN (\textbf{IKCNN}). 
\begin{table}[htb]
\centering
\caption{Accuracies on Yale}
\label{table:accuracies}
\begin{tabular}{|c|c|c|}
\hline
          & Training Set   & Validation Set  \\ \hline
VCNN & 93.3\% & \textbf{77.8}\% \\ \hline
IKCNN   & 86.7\% & \textbf{77.8}\% \\ \hline
\end{tabular}
\end{table}
With reference to table \ref{table:accuracies}, IKCNN provides equal validation accuracy to that of the VCNN despite the IKCNN having a lower training accuracy compared to the VCNN. Thus encoding the symmetry has improved the generalization of the network. Since the dataset exhibits Y-reflection symmetry, we test whether the predicted class of an image changes when it is evaluated by either network after flipping. This test is repeated for the entire training/validation set, to obtain the \emph{flip violation}. 
\begin{figure}[h]
    \centering
    \caption{Example of flip-violation}
    \includegraphics[width=.8\linewidth]{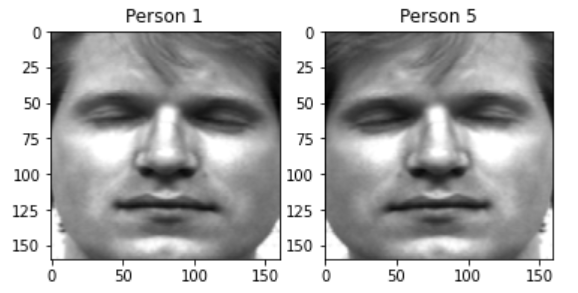}
    \label{fig:face.vio}
\end{figure}
Fig. \ref{fig:face.vio} shows an example of a flip violation. An image of person 1, when flipped, is predicted to be that of person 5 by the VCNN. Owing to its construction, IKCNN enjoys \emph{exactly} zero flip-violation. On the other hand, VCNN has a sizeable violation indicating that the notion of vertical symmetry isn't learnt by the network over the epochs of training. The flip violations are shown in table \ref{table:face.vio}.

It could be argued that this comparison is not fair as the IKCNN has a notion of symmetry and the VCNN does not. Hence, for a fairer comparison, we use data augmentation with random horizontal flips while training the VCNN to allow it to learn the inherent horizontal symmetry within the dataset. However, using data augmentation does not reduce the flip violation to the level of IKCNN. A flip violation $>10\%$ on both the training and validation sets is high on absolute terms.
\begin{table}[htb]
\centering
\caption{Flip violation}
\begin{tabular}{|c|c|c|}
\hline
                & Training Set   & Validation Set \\ \hline
VCNN & 40.8\% & 48.9\% \\ \hline
VCNN (with augmentation) & 14.17\% & 31.11\% \\ \hline
IKCNN & \textbf{0.0}\% & \textbf{0.0}\% \\ \hline
\end{tabular}
\label{table:face.vio}
\end{table}

\begin{figure}[htb]
    \centering
    \caption{Training on Yale-B}
    \begin{subfigure}{0.25\textwidth}
        \centering
        \includegraphics[width=0.8\linewidth]{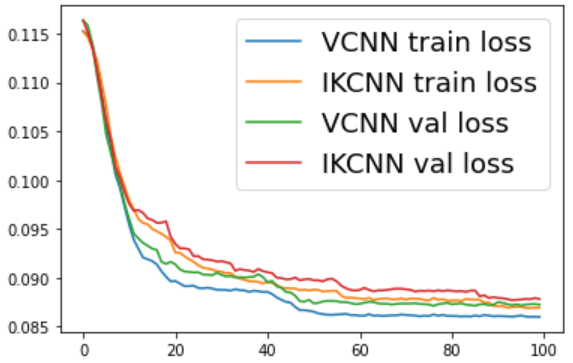}
        \caption{Loss curve}
        \label{fig:yaleb.loss}
    \end{subfigure}%
    \begin{subfigure}{0.25\textwidth}
        \includegraphics[width=0.8\linewidth]{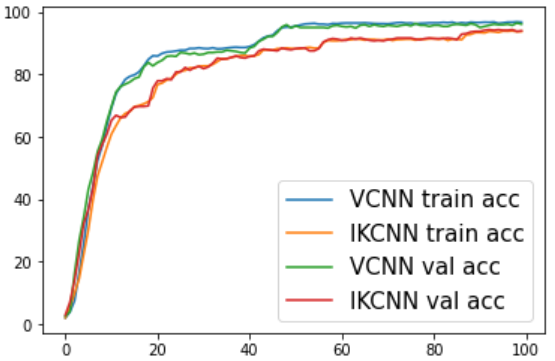}
        \caption{Accuracy curve}
        \label{fig:yaleb.acc}
    \end{subfigure}
    \label{fig:yaleb}
\end{figure}

The Yale-B \cite{yale.face} is an alternate version of the Yale dataset with 2500 images across 38 classes. It is \emph{harder} as an image task due to non-uniform lighting conditions across the images. To tackle the issue, we tested VCNN and IKCNN architectures with dropout\footnote{Using dropout makes the network inherently non-determinisitc which can cause IKCNN flip violation $>0$. Hence we have not repeated the flip-violation experiment on the Yale-B dataset
}. The IKCNN fares \emph{almost} as well as the VCNN in terms of accuracies (see table \ref{table:dropout}) reaching $>90\%$ accuracy on both training and validation sets. The training curves for this dataset are shown in Fig. \ref{fig:yaleb}. 
\begin{table}[htb]
\centering
\caption{Accuracies on Yale-B}
\label{table:dropout}
\begin{tabular}{|c|c|c|}
\hline
      & Train Accuracy & Validation Accuracy \\ \hline
VCNN  & 96.7\%         & 96.0\%              \\ \hline
IKCNN & 93.7\%         & 93.9\%              \\ \hline
\end{tabular}
\end{table}

\section{Discussion, Limitations and Impact}
\label{sec:discuss}

The techniques described in this work are only directly applicable in scenarios where an involutory invariance is inherent in the learning problem. As proof of concept, we have tested mathematical toy functions with synthetic data and obtained experimental results that match exactly with the theory. Niche applications such as data-driven models in physics \cite{pinn} have been tested as well. It must also be noted that large datasets like CIFAR-100 \cite{cifar100} are not \emph{truly} amenable to our techniques as a common underlying symmetry is not consistently exhibited across all classes. For example, the "cloud" and "sea" classes do not have any consistently represented inherent symmetry. 

There exist more general works on NN invariances such as \cite{e2cnn} which are complex and valuable in their own right. Our work, on the other hand, sets about encoding symmetries in an \emph{ab-initio} method with simple ideas that are provably exact. 

\section{Future Works}
\label{sec:future}

The discrete rotation matrix $R_{\theta}$ is also a physically relevant transformation in 2 dimensions. Making a neural network invariant to the action of discrete rotation is a physically relevant problem that should be tackled. Here $\theta=\frac{2\pi}{k}$ for an integer $k$. Differential equations exhibit a rich set of symmetries and invariances. Using \cite{neurodiffeq}, an architecture could be built to solve ODEs/PDEs, complex Hamiltonian systems that obey certain symmetries. It would also be interesting to build networks suitable to \emph{non-linear} involutory invariances. 

Although the IKCNN performs better than the VCNN aided with data-augmentation, it must be investigated how invariant kernels can act as a substitute for augmentation. This would reduce memory overheads in the training process whilst guaranteeing the underlying symmetry. The application of our techniques to other Physics tasks should be explored. Photometric redshift prediction \cite{redshift}, variable star classification \cite{varstar}, and detection of rare galaxies with symmetric features \cite{double.nuclei} are possible candidate tasks from astronomy that could see potential application. 

\section{Conclusion}
\label{sec:conclusion}

We have developed the mathematical background, theorems, and architectures to learn functions that obey involutory invariances exactly. The tests done on proposed models indicate \emph{exact} conformity to the underlying symmetry.


\end{document}